\documentclass{article}

\usepackage{arxiv}
\usepackage[numbers]{natbib}
\usepackage{amsmath}
\usepackage[utf8]{inputenc} % allow utf-8 input
\usepackage[T1]{fontenc}    % use 8-bit T1 fonts
\usepackage{hyperref}       % hyperlinks
\usepackage{url}            % simple URL typesetting
\usepackage{booktabs}       % professional-quality tables
\usepackage{amsfonts}       % blackboard math symbols
\usepackage{nicefrac}       % compact symbols for 1/2, etc.
\usepackage{microtype}      % microtypography
\usepackage{lipsum}
\usepackage{graphicx}
\graphicspath{ {./images/} }
\usepackage{cleveref}
\usepackage{listings}
\usepackage{xspace}
\usepackage{graphicx}
\usepackage{multirow}
\usepackage{enumitem}
\usepackage{booktabs} 
\usepackage{wrapfig}
\usepackage{widetable}
\usepackage{caption}

\newcommand\ourdata{ProactiveBench\xspace}

\newcommand\datasetsize{$6,790$\xspace}
\newcommand\rmperf{$91.80\%$\xspace}

\newcommand\agentperf{$66.47\%$\xspace}
\newcommand\repourl{\url{https://github.com/thunlp/ProactiveAgent}}
\title{Proactive Agent: Shifting LLM Agents from Reactive Responses to Active Assistance}

\author{
    \textbf{Yaxi Lu$^1$, Shenzhi Yang$^2$, Cheng Qian$^1$, Guirong Chen$^2$, Qinyu Luo$^1$, Yesai Wu$^1$, Huadong Wang$^1$,}\\
    \textbf{
     Xin Cong$^1$,Zhong Zhang$^1$, Yankai Lin$^2$, Weiwen Liu$^3$,
    Yasheng Wang$^3$, }\\
    \textbf{ Zhiyuan Liu$^1$, Fangming Liu$^4$, Maosong Sun$^1$}\\
    $^1$ Department of Computer Science and Technology, Tsinghua University \\
    $^2$ Gaoling School of Artificial Intelligence, Renmin University of China \\
    $^3$ Huawei Noah’s Ark Lab, 
    $^4$ Peng Cheng Laboratory \\
    \texttt{lyx23@mails.tsinghua.edu.cn, mrlyk423@gmail.com, liuzy@tsinghua.edu.cn}
}

\begin{document}
\maketitle
% \begin{minipage}{\textwidth}
% \maketitle
% \end{minipage}

\begin{abstract}
Agents powered by large language models have shown remarkable abilities in solving complex tasks. However, most agent systems remain reactive, limiting their effectiveness in scenarios requiring foresight and autonomous decision-making. In this paper, we tackle the challenge of developing proactive agents capable of anticipating and initiating tasks without explicit human instructions.
We propose a novel data-driven approach for this problem. Firstly, we collect real-world human activities to generate proactive task predictions. These predictions are then labeled by human annotators as either accepted or rejected. The labeled data is used to train a reward model that simulates human judgment and serves as an automatic evaluator of the proactiveness of LLM agents.
Building on this, we develop a comprehensive data generation pipeline to create a diverse dataset, \ourdata, containing \datasetsize events. 
% Our experiments demonstrate that the reward model achieves an F1-Score of \rmperf on our human-annotated test set. 
Finally, we demonstrate that fine-tuning models with the proposed \ourdata can significantly elicit the proactiveness of LLM agents.
Experimental results show that our fine-tuned model achieves an F1-Score of \agentperf in proactively offering assistance, outperforming all open-source and close-source models.
These results highlight the potential of our method in creating more proactive and effective agent systems, paving the way for future advancements in human-agent collaboration.
\footnote{Our code and data are available at~\repourl.}

% Agents powered by large language models have demonstrated remarkable capabilities in solving complex tasks. 
% However, most agent systems remain reactive and can not proactively assist humans, limiting their effectiveness in scenarios that require foresight and autonomous decision-making. 
% This paper proposes a novel formalization for developing proactive agents using a data-driven approach.
% Specifically, we collect real-world human activities on tasks alongside proactive agents' predictions of possible tasks. 
% These predictions are labeled based on whether they are accepted by human annotators. 
% We then use these labeled predictions to train a reward model based on LLaMA-3.1-8B-Instruct, simulating human judgment.
% Finally, we develop a data generation pipeline to obtain diverse human activities and proactive predictions, resulting in a dataset, \ourdata, containing \datasetsize events.
% Our experiments show that the reward model achieves a \rmperf F1-Score on our human annotated test set. 
% We also fine-tune models and achieve a \agentperf F1-Score on the \ourdata.
% These results demonstrate the potential of our approach in creating more effective and proactive agent systems, paving the way for future advancements in human-agent collaboration.
% The code and dataset will be released af÷ter the blind review process.
\end{abstract}
\section{Introduction}

\begin{figure*}[htb]
    \centering
    \includegraphics[width=0.85\linewidth]{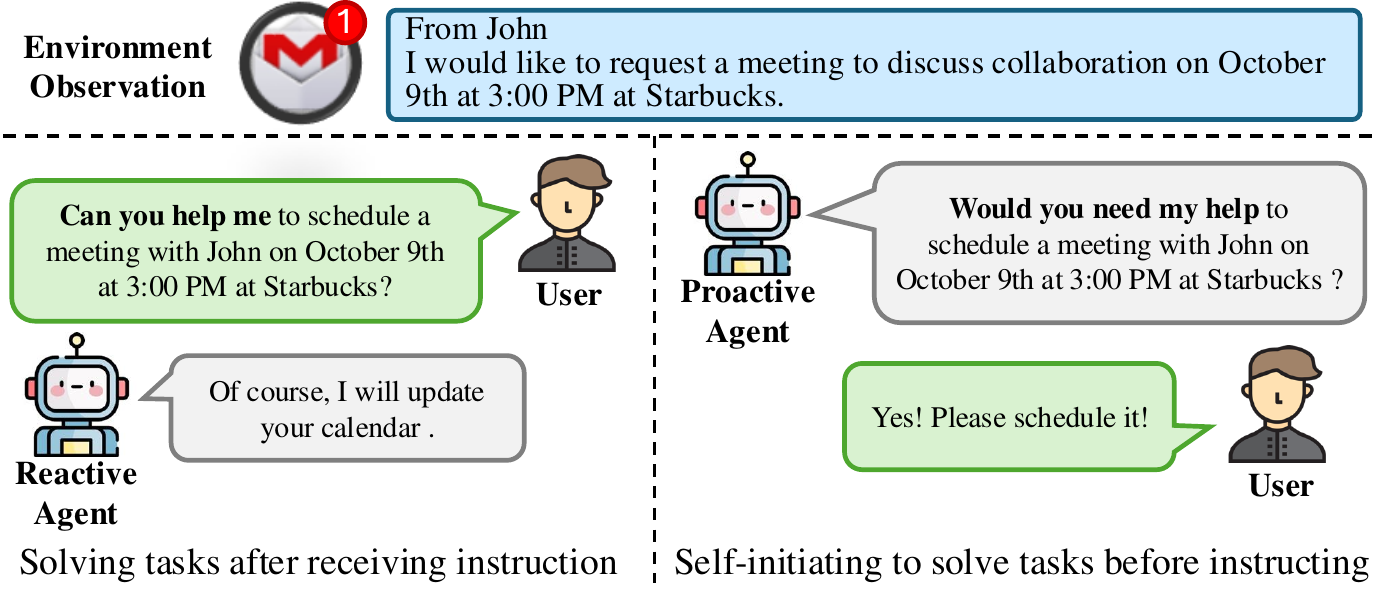}
    \caption{Comparison of agent systems with two types of human-agent interaction. Reactive agents passively receive user queries and then generate responses. Proactive agents infer tasks based on environmental observations and propose possible assistance requests accordingly.}
    \label{fig:proactive agent}
\end{figure*}

The emergence of large language models (LLMs) like ChatGPT~\citep{openaichatgptblog} has significantly advanced the development of autonomous agent~\citep{chen2023agentverse,Hong_2024_CVPR,yang2023appagent,wu2023autogen}. These LLM-based agents can understand human instructions, make plans, explore environments, and utilize tools to solve complex tasks~\citep{wu2023autogen,li2023camel} and have shown substantial potential in various applications such as robotics~\citep{doi:10.1177/02783649241281508}, personal assistants~\citep{li2024personal}, and process automation~\citep{ye2023proagent}.

Currently, most existing LLM-based agents predominantly work in the reactive paradigm: they require explicit human instructions to initiate task completion and remain dormant in terms of providing services until prompted by user instructions~\citep{ouyang2022training}. 
This paradigm limits their capacity for proactive assistance and autonomous service provision in the absence of direct human instructions.
We argue that \textbf{LLM-based agents should be proactive, capable of autonomously initiating tasks by understanding and responding to their environment}.
For instance, as illustrated in \Cref{fig:proactive agent}, the reactive agent should wait for explicit instructions from the user to execute tasks such as ``show unread emails" or ``schedule a meeting with John".
In contrast, a proactive agent would predict its task automatically by noticing an email from John suggesting a meeting and automatically offering to schedule it.
This ability of context awareness~\citep{313011} enables the proactive agent to interpret signals and proactively propose and execute tasks without explicit human instructions.
Thus, it not only significantly reduces the cognitive burden on the user but also identifies latent needs not explicitly articulated by humans. 
Consequently, the proactive agent could provide more comprehensive and seamless services to the user.

\begin{wraptable}{r}{6.5cm}
\centering
\small
\begin{tabular}{lcc}
\toprule
\textbf{Subsets} & \textbf{Scenarios} & \textbf{Entries (Tr/Ts)} \\
\midrule
Agent Model & 136 & 6,790 / 233\\
\quad \textit{Coding} & 46 & 2,275\\
\quad \textit{Writing} & 46 & 2,354\\
\quad \textit{Daily Life} & 44 & 2,161 \\
\midrule
Reward Model & - & 1,640 / 120 \\
\bottomrule
\end{tabular}
\caption{Statistics of the \ourdata, which includes three distinct settings: Coding, Writing, and Daily Life. The subset for the agent model contains $6,790$ events for training and $233$ for testing.
The subset for the reward model contains $1,640$ annotated labels for training and $120$ for testing.}
\label{tab:dataset_statistics}
\end{wraptable}

In this work, we propose a novel data-driven formalization for developing a proactive agent that anticipates user needs and takes the initiative by suggesting tasks or providing information without explicit requests.
Our approach centers on constructing \ourdata, allowing us to evaluate and enhance the agent's proactive behavior.
Firstly, we collect real-world human activity data in three settings: coding, writing, and daily life.
This includes but is not limited to, user keyboard and mouse inputs, clipboard content, browser activity, etc.
Then, we build an LLM-driven gym to generate events that reflect the raw real-world contexts we collected.
We obtain a total of $233$ events across $12$ scenarios as the test set of the \ourdata.
To further refine the proactive behavior of the LLM-based agent, we construct various events and proactive tasks under synthetic contexts with the gym.
By iterative generating more events and predictions, we obtain up to \datasetsize events as the train set of the \ourdata, as shown in~\Cref{tab:dataset_statistics}.
We fine-tune the LLaMA-3.1-8B-Instruct~\citep{touvron2023llama} and the Qwen2-7B-Instruct~\citep{qwen} on this training set to refine their proactive behavior.

% To automatically evaluate the proactive capabilities of LLMs, we train a reward model based on the LLaMA-3.1-8B-Instruct model with \annotatedatasize human judgment on proactive tasks, which achieves a \rmperf F1-Score in our evaluation.
To automatically evaluate the proactiveness of LLMs, we train a reward model that achieves up to \rmperf consistency with human judgments in terms of F1-Score, serving as an evaluator.
Using the reward model, we compared the performance of different language models on \ourdata. The results indicate that even the latest open-source models struggle to effectively predict proactive tasks. For instance, the LLaMA-3.1-8B-Instruct model only achieved a $55.06\%$ F1-Score on \ourdata.
In contrast, our fine-tuned model demonstrated significant improvements, achieving a $66.25\%$ F1-Score.
Besides, our fine-tuned Qwen2-7B-Instruct model achieves \agentperf F1-Score, outperforming all existing open-source and closed-source LLMs. 
This underscores the effectiveness of our data-driven approach in developing proactive agents, highlighting their potential to enhance user experiences across various applications.

% and evaluate them on the \ourdata.
% During the construction, we utilize the reward model to provide feedback to the proactive agent, which helps to improve the quality of the predictions.
% Next, we evaluate the proactiveness of various language models in predicting tasks the user might assign based on these events.
\section{Related Works}

% \subsection{LLM-Based Agent System}
Recent advancement in large language models~\citep{openai2023gpt4,chowdhery2022palm,touvron2023llama,zeng2022glm} has shown great progress in complex reasoning, task planning~\citep{wei2023chainofthought,gao2023pal,yao2022react,yao2023tree,liu2023llm+,ye2023large,ye2023proagent}, tool utilization~\citep{schick2023toolformer,qin2023tool,qin2023toolllm,qian2023toolink}, etc.
Consequently, a growing number of agent systems have been developed to utilize these models to tackle diverse tasks like automatic web search~\citep{qin2023webcpm}, software development~\citep{qian2023communicative,chen2023agentverse}, behavior simulation~\citep{park2023generative}.
Despite these advancements, most current agents remain predominantly reactive, passively following human instructions without sufficient context awareness~\citep{313011} to proactively meet user needs.
These reactive agents typically wait for explicit user commands, which can lead to inefficiencies as task complexity increases. 
As a result, users must constantly provide specific inputs, hindering the flow of interaction. 
In response, several works have attempted to improve the proactivity of agents. 
For example, Xuan~\citep{zhang2024ask} proposes proactive agent planning, where agents refine their tasks by actively seeking information to better understand user intentions.
Cheng Kuang and Zhi Rui~\citep{wu-etal-2024-need} study how to propose proactive support in text-to-sql and propose a novel metrics Area
 Under Delta-Burden Curve (AUDBC).
Other studies~\citep{10.1007/978-3-030-78462-1_13,10.1007/978-3-030-60128-7_2,Wiethof2021ImplementingAI,qian2024tell} focus on enabling multi-turn interactions to clarify ambiguous user instructions, which further increased cognitive load for the user.
However, these works still require the user to give an initial query before interacting with the agent.
Our approach takes a different direction by focusing on anticipating potential tasks based on monitoring user activities and environmental states, which allows the agent to proactively initialize the interaction and provide assistance.

To clarify, there are also previous works~\citep{deng2023survey} that use the term ``Proactive Agent'' to describe their dialogue systems.
However, most of these efforts~\citep{bi2021asking,ren2021learning,li2022chua,liu2024compeer} aim to enhance the helpfulness or quality of responses in a proactive manner, which differs from our focus on task anticipation and initiation.

% \subsection{Human-Agent Interaction}
% The advent of large language models~(LLMs) have led to their wide incorporation into numerous agent systems~\citep{wu2023autogen,shen2023hugginggpt,hong2023cogagent,yang2023appagent,chen2023agentverse,ye2023proagent,qian2023creator,qian2024investigate}, enhancing capabilities such as complex reasoning and human-like interaction. However, these agents often passively accept human instructions, leading to misunderstandings in certain cases due to their lack of proactive communication with humans.
% Several works~\citep{qian2024tell,10.1007/978-3-030-78462-1_13,10.1007/978-3-030-60128-7_2,Wiethof2021ImplementingAI} have attempted to improve human-agent interaction~(HAI) by enabling multi-turn interactions, aiming to better align with human intents. 
% % 修改下，不用讲过往研究的问题，对比下就行
% Our approach distinguishes itself by honing in on environments characterized by intensive interactions and high cognitive demands.
% Instead of interrupting humans and relying on explicit human-agent interactions, we try to predict human intents directly from human activities.
% This strategy minimizes the need for explicit alignment between humans and agents, offering a more streamlined and practical solution for settings where efficient interaction is paramount.

\section{Methodology} \label{method:data}

\subsection{Task Definition}
\label{method:task_def}

In our proposed proactive agent, which is distinct from traditional agent systems powered by large language models that rely on explicit user instructions, we investigate a new scenario where the agent autonomously predicts tasks users might assign, aiming to offer assistance proactively, as depicted in ~\Cref{fig:proactive agent}.
The proactive agent's mission is to give predictions based on the user's activities $A_t$, environmental events $E_t$, and state $S_t$, which can be formalized as:
\begin{equation}
P_t = f_{\theta}(E_t, A_t, S_t),
\end{equation}
where $f_{\theta}$ represents the proactive agent, parameterized by $\theta$, and $P_t$ denotes the prediction about possible task at time $t$.
It should be noticed that the prediction $P_t$ can be the predicted task or nothing if the agent believes that no task is needed.
Specifically, user activities $A_t$ contain the user's interactions with the environment and the agent, like keyboard input or chatting with the agent.
Environmental events $E_t$ contain the event that the proactive agent captured, ranging from receiving a new email to an application closed.
Environmental state $S_t$ represents the state of the current environment, like the file system state or the content of opened web pages.

In our proactive agent framework, the objective is to maximize the user's acceptance rate of the proposed tasks. 
Given the user's historical activities $A_t$, current environmental state $S_t$, and the prediction proposed by the proactive agent $P_t$, the user makes a binary decision:
\begin{equation}
    R_t = g(P_t, A_t, S_t),
\end{equation}
where $R_t$ is a binary variable indicating acceptance ($R_t = 1$) or rejection ($R_t = 0$) of the prediction.
To unify the handling of cases where the prediction $P_t$ contains no task and where it contains a task, we introduce an auxiliary variable $N_t$ that indicates whether the user needs assistance:

\begin{itemize}[nolistsep,]
    \item \( N_t = 1 \) if the user needs assistance.
    \item \( N_t = 0 \) if the user does not need assistance.
\end{itemize}

The user's acceptance \( R_t \) is then defined as:

\[
R_t = \begin{cases} 
1 & \text{if } (P_t \neq \emptyset \text{ and user accepts } P_t) \text{ or } (P_t = \emptyset \text{ and } N_t = 0) \\
0 & \text{otherwise}
\end{cases}.
\]

In this way, if the prediction \( P_t \) contains no task (i.e., the agent believes the user does not need assistance), we check the user's actual need for assistance \( N_t \). If the user indeed does not require assistance (\( N_t = 0 \)), this is marked as acceptance (\( R_t = 1 \)). Conversely, if the user requires assistance (\( N_t = 1 \)), this is marked as rejection (\( R_t = 0 \)).
Our proactive agent aims to maximize the expected acceptance rate of the proposed tasks:
\begin{equation}
\max_{\theta} \mathbb{E}[R_t].
\end{equation}

\subsection{Pipeline Overview} \label{method:pipeline}

\begin{figure*}[htb]
    \centering
    \includegraphics[width=0.96\textwidth]{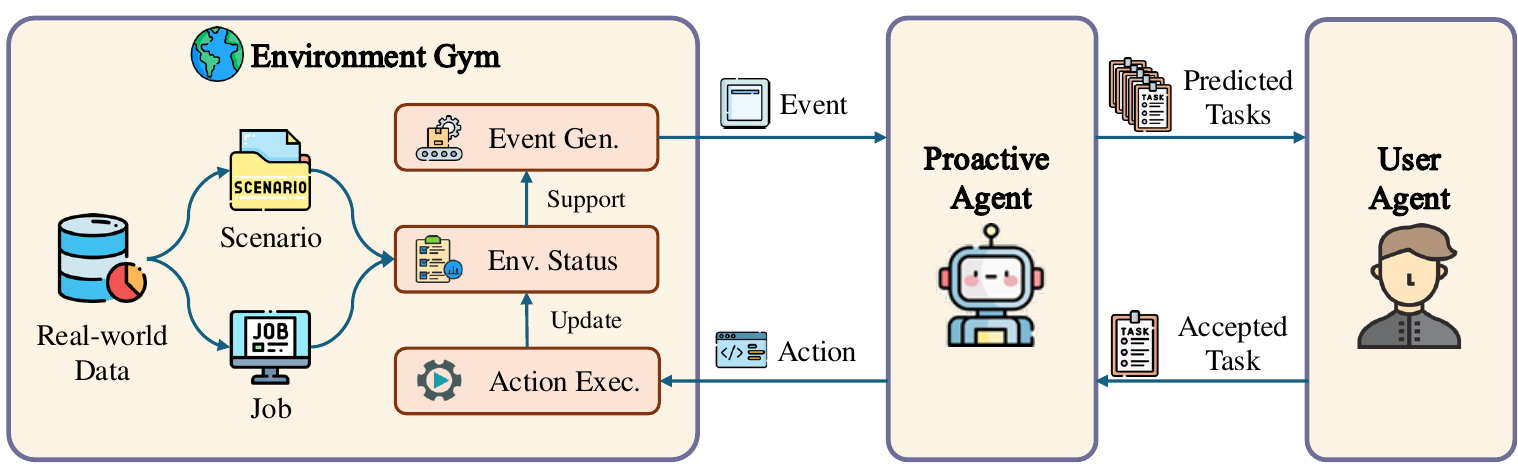}
    \caption{Overview of the data generation process. Taking daily life as an example, this process includes modules such as the initial scenario and job setup, events generation, proactive prediction, user judgment, and action execution.}
    \label{fig:simulator pipeline}
\end{figure*}

To enhance the proactive capabilities of our large language model-powered agent, we adopt a data-driven approach by building an automatic data generation pipeline. 
This pipeline simulates user activities and responses to the tasks predicted by the proactive agent across various scenarios. 
Once a prediction is accepted, we simulate the agent performing the task by interactively generating new events within the simulated environment. 
Subsequently, new user activities are created based on historical events, allowing the proactive agent to generate further predictions. 
Through this pipeline, models can learn when to generate predictions and which predictions are likely to be accepted by users. 
Specifically, our pipeline consists of three components:

(1) \textbf{Environment Gym}: This component simulates events within a specified background setting and example events, providing a sandbox for proactive agents to interact. It has two key functionalities: 
(i) Event Generation: creating potential sequences of environmental events tailored to specific scenarios;  
(ii) State Maintenance: updating and maintaining the environment’s state when new user activities are generated or when the agent performs actions during task execution.

(2) \textbf{Proactive Agent}: This component is responsible for predicting tasks that the user might assign to the agent based on the user's needs as inferred from the event history. It also interacts with tools to complete specific tasks assigned by the user.

(3) \textbf{User Agent}: This component simulates the user's activities and responses based on predefined user characteristics. It decides whether to accept and execute the tasks proposed by the agent.

In the following sections, we introduce the details of each component.

\subsection{Environment Gym}

\paragraph{Event Collection}
To improve the quality of events generated by the environment gym, we first collect real-world events as reference.
We developed a monitor software based on Activity Watcher\footnote{\url{https://github.com/ActivityWatch/activitywatch}}, which allows us to capture the details of user interactions with computer systems, including keyboard and mouse operations, visited web pages, and used development tools. 
To enhance the semantic richness of the collected data and facilitate parsing by large language models, we further merge the raw data into logically coherent segments. 
Additionally, we utilize a language model to translate the structured data into more natural textual descriptions. 
This process not only improves the interpretability of the data but also makes it more suitable for subsequent usage.

\paragraph{Scenario Generation}
After collecting reference events, rather than directly generating specific events, we generate a realistic interactive scenario to provide sufficient background information first for further generation.
To build such scenarios, we first prompt GPT-4o~\citep{OpenAIGPT4o} with the seed jobs collected from human annotators to create various jobs the user might perform under a specific category, like coding, writing, or daily life.
Then, we generate all possible entities that the tasks might involve, e.g. browser, software, and tools for the agent to perform tasks.
Next, we refine the scenario by adding more details like entity status or date time to improve the details.
Finally, the collected events are also provided to generate example events under each particular context for future events generation.
This allows us to control the granularity of events that will be generated and maintain the diversity of the scenarios.
See~\Cref{apdx:gym_prompt} for the specific prompt used.

\paragraph{Event Generation}
When it comes to specific event generation, we start with user activity generation.
For each scenario, the user agent is first requested to describe its activities and actions $A_t$ at time $t$ to complete the job in the simulated environment.
Then, the gym accepts the user's activities and actions to generate detailed events one by one.
As depicted in ~\Cref{fig:simulator pipeline}, the gym is tasked to generate logically correct and fluent events according to historical events and the current environmental state.
The key to improving the realities of the events generated and adapting to different environments is utilizing the example events we generated based on collected events during scenario generation.
Before generating events, we randomly sample the generated example events for the specific scenario and request the gym to produce new events according to them.
Once a new event $E_{t+1}$ is generated, the gym updates the entities' status in the environments and repeats the process until there are no events that can be generated with the provided user activities.
This comprehensive approach ensures that each subsequent event is not only appropriate but also contributes to a coherent and logical progression within the scenario.

\paragraph{State Maintenance}
Another important functionality of the gym is maintaining the state of the environment $S_t$.
During the scenario generation, the gym creates entities like browsers or development kits in the simulated environment, where each entity has its state and properties like the application version or the specific browser name.
When a new event is generated, the gym should update the states and properties of them to provide feedback for further event generation.
Specifically, we first gather historical state changes of related entities and prompt the GPT-4o to generate new states of the entities $S_{t+1}$ with the new event.
During the process, the simulated time will also be updated according to the granularity of the event.
After that, the next event will be generated based on the latest environment state $S_{t+1}$.

\subsection{Proactive Agent}
\label{method:agent_design}
\begin{wrapfigure}{r}{7.6cm}
    \centering
    \includegraphics[width=\linewidth]{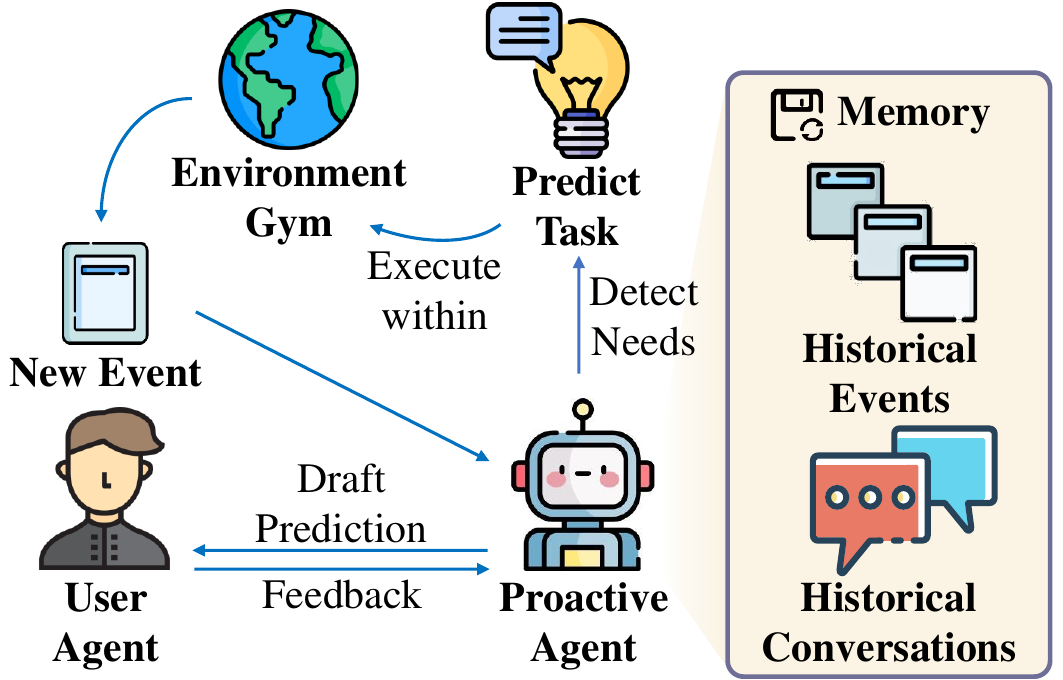}
    \caption{Overview of the proactive agent framework. The agent monitors new events and updates its memory to predict potential tasks.}
    \vspace{-30pt}
    \label{fig:agent workflow}
\end{wrapfigure}
The second component in our data generation pipeline is the proactive agent that predicts tasks the user might assign.
As detailed in ~\Cref{fig:agent workflow}, upon the agent receiving new event $E_{t}$ at time $t$, it first updates its memory with the event.
To improve the quality of the prediction, it also accepts feedback from the user agent on its draft prediction.
Combining new events with historical ones and conversations with the user, the agent incorporates user characteristics to raise potential tasks.
If the agent detects potential tasks, it will raise the task as a new event and wait for the judgment of the user agent.
Otherwise, the proactive agent predicts no potential tasks and stays silent.
Once the predicted task is accepted, the agent will execute the task within the gym, which generates multiple events about the interaction between the agent and the environment.  
During the data generation, the agent would constantly receive events from the Gym and predict potential tasks.

\paragraph{Task Execution}
As mentioned before, the proactive agent executes the predicted task once the user accepts.
This process is mainly done through multi-turn interaction between the proactive agent and the gym.
Specifically, the proactive agent will be provided with the tools generated during the scenario generation, like file system tools in the computer or access to the smart light switch, to interact with the simulated environment.
Each time the proactive agent takes an action, the gym will generate a new event, which is further processed by the gym and the user agent to update the environment state.
After that, the proactive agent detects the new environment state $S_{t+1}$ and takes new actions according to the events generated by the gym.
This process ends when the user interrupts or the proactive agent finishes its tasks.

\subsection{User Agent}
\label{method:annotation}

The user agent is designed to emulate users' activities $A_t$ and responses about the agent's prediction $P_t$, as illustrated in ~\Cref{fig:simulator pipeline}. 
We prompt GPT-4o to generate activities and actions for the provided task in the specific environment.
The gym further processes the activities and actions to generate a new event.
Then the proactive agent predicts potential tasks according to the events.
Upon receiving the predicted task, the user agent determines whether to accept or reject it. 
If the user agent accepts the task, the proactive agent will set up and execute the accepted task within the environment gym. 
Otherwise, if the user agent declines the suggested assistance, the environment gym generates new events autonomously without any interventions.
In our settings, we collect judgment from human annotators and train a reward model to simulate the judgment.

Specifically, to ensure the reward model aligns closely with human judgment, we generate and annotate a dedicated dataset to indicate whether humans would accept the predicted task or not. 
We utilize 9 different language models to generate diverse predictions for each event.
Between these predictions, we select 5 predictions with minimum total cosine distance as our label target.
Each prediction is annotated with one of three options by three separate annotators: \textit{accept}, \textit{reject}, or \textit{reject all}. 
The \textit{reject all} option is chosen when the annotator believes that the given events did not imply any possible tasks that the user might assign, aka $N_t = 0$ in~\Cref{method:task_def}.
Otherwise, if one prediction is labeled as accepted, we label the event $E_t$ with $N_t = 1$.
We use majority voting to make the final decision on each prediction.
After all, the annotation results in a dataset of $1,760$ entries, each containing event traces, task predictions, and decisions on accepting the predicted task from three distinct annotators.
Notably, our annotators achieve an impressive consistency rate of over $91.67\%$ on the test set, underscoring the annotations' reliability and the dataset's robustness for further analysis.
To further facilitate automatic data generation, we also prompt the GPT-4o to produce a more detailed explanation of the user judgment.
Further details regarding the assessment of the reward model are available in~\Cref{exp:reward_model_assessment}.

\section{Experiments}\label{sec:exp}

\subsection{Reward Model Assessment}
\label{exp:reward_model_assessment}

To automatically evaluate whether the predicted tasks and their timing are appropriate, we seek to train a reward model capable of accurately imitating user judgments. To this end, we apply the user-annotated data to train LLaMA-3.1-8B-Instruct~\citep{touvron2023llama} and compare it with several baselines to show its superiority.

\paragraph{Setting.} We use the $1,760$ entries with human annotations and randomly split them into a training set ($1,640$ entries) and a test set ($120$ entries). 
We then train LLaMA-3.1-8B-Instruct on the training set to obtain our reward model. 
We employ a total batch size of $32$, a learning rate of $1e-5$, and an Adam Optimizer with a $0.01$ warm-up ratio. 
We train the reward model for $3$ epochs to prevent it from over-fitting. We use 8 A100 GPUs on one node to train for approximately 1.5 hours.
The detailed prompt template is listed in~\Cref{apdx:evaluator}. 
We use the test split to evaluate our adapted model and all the baselines. 
To be noticed, our human annotators achieve up to $91.67\%$ agreement ratio on the test set, demonstrating the effectiveness of our evaluation.

\paragraph{Metrics.}
% During the evaluation, we prompt models to generate acceptance of each prediction and compare it with human-annotated results.
We use the reward model to perform binary classification on whether to accept predicted tasks and compare its results with human-annotated results.
This assesses how well the reward model aligns with human judgment regarding the suitability of the predicted tasks.
% This evaluates whether models could successfully align with human judgments regarding whether the proposed agent task is appropriate. 
We compare the judgments made by the reward models and humans to calculate the Recall, Precision, Accuracy, and F1-Score.
Additionally, we calculate the agreement ratio for the following cases: 
% We test the agreement between the reward model and human annotators in the following scenarios:
\begin{itemize}[nolistsep,noitemsep,left=0pt]
\item \textbf{Missed-Needed}: $N_t=1, P_t=\emptyset$, the user needs help, but the agent does not provide it.
\item \textbf{Non-Response}: $N_t=0, P_t = \emptyset$, the user does not need help, the agent does not prompt.
\item \textbf{Correct-Detection}: $N_t=1, P_t \neq \emptyset$, and the user accepts the task predicted by the agent.
\item \textbf{False-Detection}: $N_t=0, P_t \neq \emptyset$, the user does not need help but agent prompts.
\end{itemize}
% For each scenario, we compare the judgment between the reward models and humans to obtain the Recall, Precision, Accuracy, and F1-Score in~\Cref{tab:reward_models}.

\begin{table}
% \small
\centering
\begin{widetabular}{\textwidth}{lccccc}
\toprule
& GPT-4o  & GPT-4o-mini & LLaMA-3.1-8B & LLaMA-3.1-70B  & Ours    \\
\midrule
Agree. MN$^{\uparrow}$        & 3.33\%  & 56.67\%     & \textbf{80.00\%}      & 33.33\%        & \textbf{80.00\%}  \\
Agree. NR$^{\uparrow}$       & \textbf{100.00\%} & 56.67\%     & 30.00\%      & 83.33\%        & 86.67\%  \\
Agree. CD$^{\uparrow}$  & \textbf{100.00\%} & 86.67\%     & 96.67\%      & \textbf{100.00\%}       & \textbf{100.00\%} \\
Agree. FD$^{\uparrow}$        & 0.00\%  & 33.33\%     & 13.33\%      & 6.67\%         & \textbf{100.00\%} \\
\midrule
Recall$^{\uparrow}$                  & \textbf{100.00\%}     & 71.67\%     & 63.33\%      & 91.67\%        & 93.33\%  \\
Precision$^{\uparrow}$               & 50.42\% & 56.58\%     & 54.29\%      & 53.40\%        & \textbf{90.32\%}  \\
Accuracy$^{\uparrow}$                & 50.83\% & 58.33\%     & 55.00\%      & 55.83\%        & \textbf{91.67\%}  \\
\midrule
F1-Score$^{\uparrow}$                & 67.04\% & 63.24\%     & 58.46\%      & 67.48\%        & \textbf{91.80\%}  \\
\bottomrule
\end{widetabular}

\caption{Evaluation results of different models on our test set as the reward model.
For the Missed-Need (MN), Non-Response (NR), Correct-Detection (CD), and False-Detection (FD) scenarios, we present the \textbf{agreement ratio} between models and the major voting of our human annotators.
Our model which is fine-tuned based on LLaMA-3.1-Instruct-8B achieves the best F1-Score of \rmperf.}\label{tab:reward_models}
\end{table}

\paragraph{Results.}
As~\Cref{tab:reward_models} shows, all existing models perform well on correct detection, but perform badly in other scenarios, especially in the false alarm scenario.
After a deeper analysis, we find that existing models just can not infer what the user might need and tend to accept arbitrary help, even if it is very abstract or meaningless to current observation.
In contrast, our reward model achieves a $100\%$ agreement ratio on false alarm scenario and a solid $91.80\%$ F1-Score across all scenarios.
We select our reward model for further analysis across the~\ourdata.

\subsection{Proactive Agent Evaluation}

\begin{table}[htb]
\centering
\begin{widetabular}{\textwidth}{lcccccc}
\toprule
\textbf{Model} & \textbf{Recall$^{\uparrow}$} & \textbf{Precision$^{\uparrow}$} & \textbf{Accuracy$^{\uparrow}$} & \textbf{False-Alarm$^{\downarrow}$} & \textbf{F1-Score$^{\uparrow}$} \\
\midrule
\multicolumn{6}{c}{\textit{\textbf{Proprietary models}}} \\
\midrule
Claude-3-Sonnet & 27.47\% & 37.31\% & \textbf{52.42\%} & 62.69\% & 31.65\% \\
Claude-3.5-Sonnet & 97.89\% & 45.37\% & 49.78\% & 54.63\% & 62.00\% \\
GPT-4o-mini & \textbf{100.00\%} & 35.28\% & 36.12\% & 64.73\% & 52.15\% \\
GPT-4o & 98.11\% & \textbf{48.15\%} & 49.78\% & \textbf{51.85\%} & \textbf{64.60\%} \\
\midrule
\multicolumn{6}{c}{\textit{\textbf{Open-source models}}} \\
\midrule
LLaMA-3.1-8B & 98.86\% & 38.16\% & 39.06\% & 61.84\% & 55.06\% \\
LLaMA-3.1-8B-Proactive & \textbf{99.06\%} & \textbf{49.76\%} & \textbf{52.86\%} & \textbf{50.24\%} & \textbf{66.25\%} \\
\midrule
Qwen2-7B & 98.02\% & 44.00\% & 43.61\% & 56.00\% & 60.74\% \\
Qwen2-7B-Proactive & \textbf{100.00\%} & \textbf{49.78\%} & \textbf{50.66\%} & \textbf{50.22\%} & \textbf{66.47\%} \\
\bottomrule
\end{widetabular}
\caption{Evaluation results of different models' performance on the~\ourdata. The GPT-4o stands out for close-sourced models, achieving over $64.60\%$ F1-Score. For open-sourced models, our fine-tuned Qwen2-7B model achieves best result, with a~\agentperf F1-Score.}
\label{tab:evaluation_result}
\end{table}

\paragraph{Setting.}
We use the training set of \ourdata to obtain the Proactive Agent based on the two open-source models: LLaMA-3.1-8B-Instruct and Qwen2-7B-Instruct.
During training, we employ a total batch size of $32$, a learning rate of $1e-5$, and an Adam Optimizer with a $0.01$ warm-up ratio. 
We train the model for $3$ epochs. 
We use 8 A100 GPUs on one node to train for approximately 2 hours.
The detailed prompt can be found in~\Cref{apdx:agent_model}.
The automatic evaluation of these metrics relies on the simulated judgment given by the reward model.
All models are evaluated in our test split of the~\ourdata, which contains $233$ events collected in the real world.
We employ the same prompt template across all models. The temperature is set to $0$ during testing.

\paragraph{Metrics.}
We evaluate the performance of the Proactive Agent based on whether the user accepts its prediction. 
As described in~\Cref{method:task_def}, the user's acceptance $R_t$ contains four conditions.
In our specific settings, \textbf{Recall} measures the proportion of actual needs for assistance that were correctly predicted by the agent, including cases where the agent predicts a task and the user accepts it, as well as cases where no task is predicted and the user does not need assistance. 
\textbf{Precision} measures the proportion of predicted tasks that were actually accepted by the user.
\textbf{Accuracy} measures the overall correctness of the agent's predictions. 
The \textbf{False-Alarm} measures the proportion of incorrect task predictions, specifically when a task is predicted but not needed. 
The \textbf{F1-Score} provides a balanced measure of the goodness of the agent's proactive behavior.
We use the reward model during the evaluation to automatically generate the user's judgment.
Based on the confusion matrix, we report Recall, Precision, Accuracy, False Alarm, and F1-Score across all settings. The detailed calculation method can be found in~\Cref{apdx:agent_model}.

\paragraph{Result.}
\Cref{tab:evaluation_result} compares various models on the test set of the~\ourdata, which contains $233$ events collected from the real world user.
Close-sourced models like GPT-4o~\citep{OpenAIGPT4o} or GPT-4o-mini all tend to predict proactive tasks actively.
Most of them succeed in assisting when the user needs but fail to stay silent when the user does not require any assistance, resulting in a relatively high false alarm ratio.
For example, the GPT-4o-mini provides unnecessary assistance even when the events provided do not contain meaningful operations, like switching between software but doing nothing else.
Another big issue is early assistance when no precise users' intents can be found in the given observation.
This makes the proactive tasks proposed by the model seem too abstract or useless, resulting in a relatively high false alarm ratio.
The Claude-3-Sonnet~\citep{Anthropic2024Claude} shows a different example of failing to detect the user's need and provide assistance that does not satisfy the user's expectation.

For open-sourced models, we evaluate the performance of the LLaMA-3.1-Instruct-8B and Qwen2-Instruct-7B before and after fine-tuning based on our synthesized data.
As shown in~\cref{tab:evaluation_result}, both models obtain an impressive improvement, especially for LLaMA-3.1-8B, which improves its F1-Score from $44.78\%$ to $61.74\%$.
The results demonstrate the effectiveness of our data synthesis pipelines.
As for the concern of being overly interrupted by the proactive agent, our fine-tuned models achieve solid progress in reducing the false alarm ratio, which is comparable to the performance of the GPT-4o.
Besides, the finetuned Qwen2-7B is also outperform the GPT-4o in terms of the F1-Score, resulting in the highest F1-Score of $66.07\%$
However, we also observed the same pattern of models tends to provide as much assistance as possible, instead of providing necessary assistance when the user needs it.

In short, while most models can assist when needed, they still frequently offer unnecessary help, even when instructed to provide only essential assistance.

\subsection{Performance Analysis}
In this section, we analyse two possible type of settings that could impact the performance of the proactive agent.

\paragraph{Predict Multiple Tasks.}
When it comes to real-world applications, the proactive agent can provide multiple candidate tasks to improve overall performance. 
To evaluate how models perform under this condition, we allow them to generate multiple candidate tasks at once, but no more than three to avoid a high cognition burden for the user.
In this setting, we let the reward model check the candidate tasks one by one.
We label the result as accepted if one of the candidate tasks is accepted, and rejected if only all the candidate tasks are rejected.

As shown in~\Cref{tab:multi}, all models obtain solid improvement across all metrics when comparing ``pred@1'' with ``pred@3''.
Take the GPT-4o as an example, it obtains higher accuracy and precision while reducing its false alarm by providing diverse candidate tasks.
The huge drops in the false alarm ratio, from $51.85\%$ to $36.44\%$ are mainly due to its improvement in providing proactive tasks.
However, when comparing GPT-4o-mini with LLaMA-3.1-8B, we observed different degrees of improvement.
These two models perform similarly when predicting only one proactive task at once, but show a nearly $9\%$ difference in terms of F1-Score when predicting multiple candidates at once.
We analyzed the result and found that the LLaMA-3.1-8B tends to provide unexpected assistance when the user's need is unclear, which can not be improved by providing multiple candidates.

\begin{table}[htb]
\centering
\begin{widetabular}{\textwidth}{lcccccc}
\toprule
\textbf{Model} & \textbf{Settings} & \textbf{Recall$^{\uparrow}$} & \textbf{Precision$^{\uparrow}$} & \textbf{Accuracy$^{\uparrow}$} & \textbf{False-Alarm$^{\downarrow}$} & \textbf{F1-Score$^{\uparrow}$} \\
\midrule
GPT-4o-mini & pred@1 & 100.00\% & 35.28\% & 36.12\% & 64.73\% & 52.15\% \\
            & pred@3 & 99.32\% & 65.32\% & 66.52\% & 34.68\% & 78.80\% \\
            & w/ RM & 55.45\% & 63.54\% & 63.95\% & 36.46\% & 59.22\% \\
            & pred@3, w/ RM & 100.00\% & 65.35\% & 66.09\% & 34.65\% & 79.05\% \\
\midrule
GPT-4o & pred@1 & 98.11\% & 48.15\% & 49.78\% & 51.85\% & 64.60\% \\
       & pred@3 & 100.00\% & 63.56\% & 64.81\% & 36.44\% & 77.72\% \\
       & w/ RM & 56.76\% & 55.26\% & 57.61\% & 44.74\% & 56.00\% \\
       & pred@3, w/ RM & 100.00\% & 63.30\% & 65.67\% & 36.70\% & 77.53\% \\
\midrule
LLaMA-3.1-8B & pred@1 & 98.86\% & 38.16\% & 39.06\% & 61.84\% & 55.06\% \\
              & pred@3 & 100.00\% & 52.79\% & 52.79\% & 47.21\% & 69.10\% \\
              & w/ RM & 77.08\% & 42.52\% & 47.64\% & 57.41\% & 54.81\% \\
              & pred@3, w/ RM & 95.12\% & 61.58\% & 66.09\% & 38.42\% & 74.76\% \\
\bottomrule
\end{widetabular}
\caption{Comparison between different settings for each model. The setting ``pred@1'' means predicting one task at a time. The setting ``pred@3'' means predicting 3 tasks at a time. The setting ``w/ RM'' means we will provide feedback from the reward model to help better prediction.}
\label{tab:multi}
\end{table}

\paragraph{Feedback From the Reward Model.}
We also investigate whether the feedback from our reward model could help models improve their performance on the~\ourdata.
This is done with the same logic as described in~\Cref{fig:agent workflow}.
For each model, we first ask them to generate a draft prediction and obtain feedback from the user agent (which is built on the reward model in this case).
Then we let the model refine its prediction to obtain the final prediction.

As shown in~\Cref{tab:multi}, by adding the feedback from the reward model (settings with ``w/ RM''),  models generally reduce their false alarm ratio and improve the accuracy, but drop dramatically in terms of the recall.
We observe that models stay silent once they receive feedback from the reward model.
However, doing nothing is not always the optimal solution. 
The GPT-4o seems to fail to capture the possible user needs, leading to a performance drop in terms of F1-Score.
For other models like GPT-4o-mini and LLaMA-3.1-8B, they deed obtain a marked improvement in terms of the F1-Score.
Another setting combining the multiple-task prediction with the reward model (``pred@3, w/ RM'') shows a more general improvement across the board.
By integrating the reward model into the Proactive Agent, we can make the Proactive Agent more smartly detect user needs and reduce the false alarm ratio even when we can not access the weight directly, which is good news for developing the Proactive Agent.

\subsection{Case Study}
In this section, we explore two prevalent types of failures encountered in predicting possible tasks: the inability to detect user needs and making predictions at inappropriate times.

\begin{figure*}[htbp]
\centering
\includegraphics[width=0.95\linewidth]{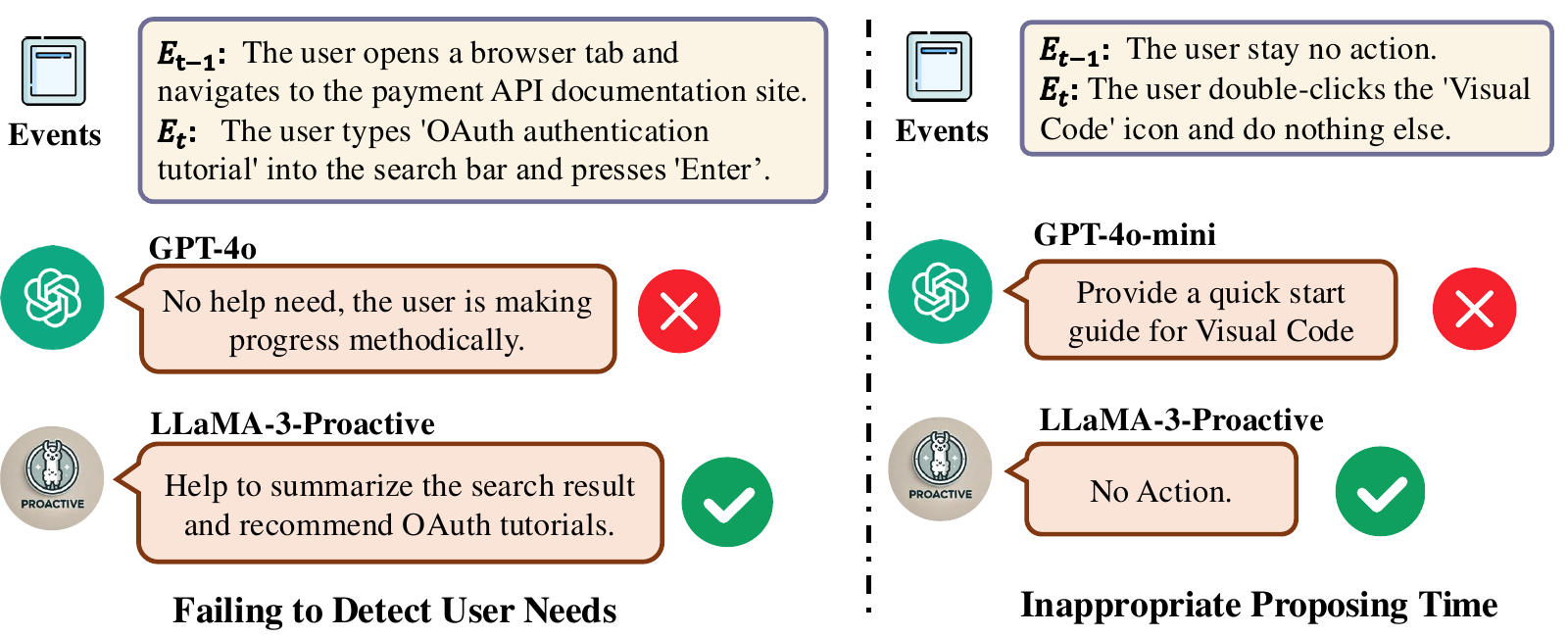}
\caption{
Two types of failure: failing to detect user needs (left) and inappropriate proposing time (right). We compare the response between our fine-tuned LLaMA-3.1-Instruct-8B with other models to show the refined proactive behavior of the model.
}
\label{fig:Failure Cases}
\end{figure*}

As illustrated in ~\Cref{fig:Failure Cases} (left), a notable failure occurs when the GPT-4o model does not assist at crucial moments. 
For instance, when a user is engaged in integrating a payment API and requires a tutorial for guidance, the model remains silent. 
Instead, our model successfully detects human needs and offers assistance.
The underlying intention to minimize disruptions ironically leads to missed opportunities to offer timely help.

Conversely, the right side of ~\Cref{fig:Failure Cases} showcases an instance of ill-timed prediction. Here, the GPT-4o-mini suggests an action when there are no user needs showing in events.
This scenario underscores the possible unintended events existing in human activities.
The model should judge whether there are possible tasks smartly to avoid unnecessary actions.
These instances highlight the intricate nature of human activities and the sophisticated reasoning required for models to accurately predict human needs. 
To navigate the delicate balance between being helpful and intrusive, models must develop a deeper understanding of user contexts and activities, ensuring their interventions are both timely and pertinent.

\section{Conclusion}
We present an innovative approach to human-agent interaction by leveraging proactive task predictions that anticipate human needs. 
We introduce \ourdata, a comprehensive dataset comprising \datasetsize events, designed to refine the proactive behavior of LLM-based agents and establish an automatic benchmark for assessing model proactiveness. 
By iteratively generating events in synthesized scenarios, we create training data that enhances the proactive capabilities of our models.
Our experiments demonstrate significant improvements in the agent's performance on \ourdata, validating the effectiveness of our methods. 
Despite these advancements, our findings underscore ongoing challenges, particularly in minimizing inappropriate task proposals and ensuring task predictions are contextually accurate.
Future research should focus on enhancing the precision and timeliness of task predictions to improve the efficacy of the proactive human-agent interaction.

% \section*{Ethics Statement}
% In the construction of our dataset, we have adhered to strict ethical standards. 
% The human annotators are students from the computer sciences area and obtain rewards for their annotations.
% We use ChatGPT in this research.
% All the original data utilized for the compilation of our dataset are either sourced from open-access platforms or have been used with the appropriate permissions granted. 
% We have conducted a thorough review to ensure that our dataset is devoid of any harmful content or private information. 
% Our commitment to ethical research practices extends to the respect for privacy, intellectual property rights, and the integrity of the data sources.

\section*{Limitations}
While our method demonstrates that it can effectively and proactively predict possible tasks, the current research is constrained by several limitations. 
Firstly, the environment settings we have explored are still limited. 
The contexts in this paper provide a foundational understanding, but broader application areas need to be investigated to fully establish the versatility and robustness of the proactive agent.
Moreover, models still exhibit a relatively high ratio of false alarms, indicating that they cannot yet perfectly predict possible tasks. 
This limitation highlights the need for further refinement of the model's proactive behavior to avoid bothering the user. 
The high rate of false alarm can lead to unnecessary or incorrect actions, which may reduce user trust and the overall efficiency of the system.
The dynamic adjustment of the proactiveness of the agent according to the specific context should be explored in more depth.
Future research should focus on several key areas to address these limitations:

\begin{itemize}[noitemsep,nolistsep,left=0pt]
    \item \textbf{Expansion of Environment Settings:} Research should explore a wider variety of scenarios and contexts to validate the model's generalizability. This includes domains where proactive prediction of tasks can significantly enhance user experience and operational efficiency.
    \item \textbf{Improvement in Prediction Accuracy:} Efforts should be directed towards reducing the false alarm rate by enhancing the model's understanding of context and user behavior. 
    \item \textbf{User-Centric Evaluation:} Future studies should involve extensive user-centric evaluations to better understand how users interact with the proactive agent and to identify areas for improvement. User feedback and behavioral data can provide valuable insights into refining the prediction algorithms and making the system more intuitive and reliable.
    \item \textbf{Ethical and Privacy Considerations:} As the proactive agent needs the environment information for prediction tasks, it is crucial to address ethical and privacy concerns. Ensuring that user data is handled responsibly and that the agent operates transparently and within ethical guidelines will be critical for gaining user trust and acceptance.
\end{itemize}

\bibliographystyle{unsrt}  
\bibliography{references}  %%% Remove comment to use the external .bib file (using bibtex).

\newpage
\appendix
\section*{Appendix}

\section{Reward Model Training Setting}
\label{apdx:evaluator}
We use Llama-3.1-Instruct-8B as the base model for our training. The total dataset size is approximately $1,640$. Specifically, we employ a total batch size of $32$, a learning rate of $1e-5$, and an Adam Optimizer with a $0.01$ warm-up ratio. We train the reward model for 3 epochs to prevent it from over-fitting. We use 8 A100 GPUs on one node to train for approximately 1.5 hours.
\noindent\makebox[\linewidth]{\rule{\linewidth}{0.4pt}}\\
\textit{Prompt Template}
\begin{lstlisting}[basicstyle=\ttfamily, breaklines=true]
<Task>
Evaluate the task proposed by the proactive assistant as the user.
</Task>

<Rule>
0. Analyze the current observation to understand your current situation and requirements.
1. If the proposed task is `null` (indicating no task is proposed under the current observation), follow these steps:
   - Accept the `null` task if you believe there is no need for a task.
   - Reject the `null` task if you believe a task is needed.
2. Minimize interruptions from the assistant by only accepting tasks that are valuable.
3. Evaluate the current observation and make a judgment on the proposed task accordingly.
</Rule>

<Format>
You should answer with the following JSON format:
{
    "thought": "Give your thoughts first, then provide the judgment of the task.",
    "judgment": "accepted or rejected"
}
</Format>
\end{lstlisting}

\section{Agent Model Training Setting}
\label{apdx:agent_model}
Similarly, we use Llama-3-Instruct 8B and Qwen2-Instruct-7B as the base model for agent model training. 
The total dataset size is approximately $6,790$. Specifically, we employ a total batch size of $32$, a learning rate of $1e-5$, and an Adam Optimizer with a $0.01$ warm-up ratio. We train the model for 3 epochs to prevent it from over-fitting. We use 8 A100 GPUs on one node to train for approximately 2 hours.

\paragraph{Template.} We apply the following prompt template to train the agent model:

\noindent\makebox[\linewidth]{\rule{\linewidth}{0.4pt}}\\
\textit{Prompt Template}
\begin{lstlisting}[basicstyle=\ttfamily, breaklines=true]
<Role> You are a helpful assistant that provides proactive suggestions to the user. </Role>

<Task> Understand what the user is doing and anticipate their needs based on events. Only propose assistance when you fully understand the user's actions. Use available operations to ensure the task is feasible. Execute the task if the user accepts your proposal. </Task>

<Format> Respond in the following JSON format:
{
    "Purpose": "The purpose of the user's last action.",
    "Thoughts": "Your thoughts on the user's actions.",
    "Proactive Task": "Describe your proposed task, or set to `null` if no assistance is needed.",
    "Response": "Inform the user about your assistance if proposing a task."
}
</Format>

<Rules>
- Ensure the proposed task is relevant to the events. - Focus on the user's current needs and predict helpful tasks.
- Consider the timing of events.
- Only offer proactive assistance when necessary.
- Deduce the user's purpose and whether they need help based on event history.
- Set `Proactive Task` to `null` if the user doesn't need help.
</Rules>
\end{lstlisting}

\section{Prompt Template for Environment Gym}
\label{apdx:gym_prompt}

\subsection{Prompt for Scene Generation}
\noindent\makebox[\linewidth]{\rule{\linewidth}{0.4pt}}\\
\textit{Prompt Template}
\begin{lstlisting}[basicstyle=\ttfamily, breaklines=true]
<Role>
You are tasked with simulating an environment within a system. The content labeled `Source: environment` reflects your past actions and decisions.
</Role>

<Task>
Generate and refine detailed environment settings. Based on the latest activities, create multiple events to describe changes in the environment.
</Task>

<Rules>
- Ensure the subject of the generated content aligns with the latest activities's source.
- Avoid subjective opinions or emotions; focus on objective changes.
- Ensure events are consistent with historical events labeled `[events]` and include all - changes from the activities.
- Introduce occasional failures or unexpected events for realism.
- Ensure each event is logically connected to the previous one and does not include nonexistent elements.
- Pay close attention to entity operations; if an operation is not allowed or impractical in the real or simulated environment, raise an error and explain the issue.
</Rules>
\end{lstlisting}

\subsection{Seed Jobs Data}
\noindent\makebox[\linewidth]{\rule{\linewidth}{0.4pt}}\\
\textit{Prompt Template}
\begin{lstlisting}[basicstyle=\ttfamily, breaklines=true]
<Task>
You are tasked to generate realistic scenarios where a user might need assistance from an AI assistant. Always remember to keep the scene realistic and believable by including as much details as possible.
</Task>

<Rule>
- You will iteratively generate more information about the scene. Make sure each time you add a new detail, it is consistent with the previous details. Always generate new content based on the previous generated content.
- You can add as many details as you want, but make sure they are consistent with the previous details.
- Try to generate diverse details about the scene. You will be tasked to simulate events in the scene later.
</Rule>
\end{lstlisting}

\subsection{Prompt for User Agent Generation}
\noindent\makebox[\linewidth]{\rule{\linewidth}{0.4pt}}\\
\textit{Prompt Template}
\begin{lstlisting}[basicstyle=\ttfamily, breaklines=true]
<Role>
You are tasked with simulating a user within a system. The content labeled `Source: user` reflects your past actions and decisions.
</Role>

<Task>
Generate human-like activities with distinct characteristics and identities. You will receive events and observations from the environment; analyze these closely to decide your actions.
</Task>

<Rules>
- Respond like a real user; don't be overly predictable.
- Refer to # User Info to understand your identity.
- Critically evaluate the received information, as it may not always be accurate.
- Stay aware of environmental changes, which can occur at any time.
</Rules>
\end{lstlisting}

\subsection{Prompt for Status Updating}
\noindent\makebox[\linewidth]{\rule{\linewidth}{0.4pt}}\\
\textit{Prompt Template}
\begin{lstlisting}[basicstyle=\ttfamily, breaklines=true]
<Task>
Evaluate the task proposed by the proactive assistant as the user.
</Task>

<Rule>
0. Analyze the current observation to understand your current situation and requirements.
1. If the proposed task is `null` (indicating no task is proposed under the current observation), follow these steps:
   - Accept the `null` task if you believe there is no need for a task.
   - Reject the `null` task if you believe a task is needed.
2. Minimize interruptions from the assistant by only accepting tasks that are valuable.
3. Evaluate the current observation and make a judgment on the proposed task accordingly.
</Rule>

<Format>
You should answer with following JSON format:
{
    "thought": "Give your thoughts first, then provide the judgement of the task.",
    "judgement": "accepted or rejected"
}
</Format>
\end{lstlisting}

\subsection{Metrics Calculation}
\paragraph{Definition}
Here is how we define the label of each prediction.
\begin{itemize}[nolistsep,left=0pt]
\item \textbf{True Positive (TP):} Agent predicts task, the user accepts.
\item \textbf{False Positive (FP):} Agent predicts task, the user rejects.
\item \textbf{True Negative (TN):} Agent does not predict a task, and the user does not need assistance.
\item \textbf{False Negative (FN):} Agent does not predict the task, but the user needs assistance ($N_t = 1$ in~\Cref{method:task_def}).
\end{itemize}

\paragraph{Recall} A high recall indicates that the agent frequently identifies situations where help is needed. This metric is crucial for assessing the agent's ability to recognize and respond to user needs on time. 
\begin{equation}
    Recall = \frac{TP}{TP+FN}
\end{equation}

\paragraph{Precision} A high precision indicates that the agent proposes good tasks while not bothering the user too much. This metric is crucial when considering the annoying behavior of the proactive agent could greatly reduce user satisfaction. 
\begin{equation}
    Precision = \frac{TP}{TP+FP}
\end{equation}

\paragraph{Accuracy} High Accuracy demonstrates that the agent has a good understanding of user needs, as most of its predictions are accepted. This metric is essential for measuring the relevance and correctness of the agent's proactive suggestions.
\begin{equation}
    Accuracy = \frac{TP+TN}{P+N}
\end{equation}

\paragraph{F1-Score} High F1-Score means the proactive agent strikes a good balance between being helpful and proactive.
\begin{equation}
    F_1 = 2*\frac{Recall * Precision}{Recall + Precision}
\end{equation}
\section{Data Examples}
\subsection{Event Samples}
\noindent\makebox[\linewidth]{\rule{\linewidth}{0.4pt}}\\
\textit{Collected Raw Data}
\begin{lstlisting}[basicstyle=\ttfamily, breaklines=true]
[{
    "timestamp": 1717335890.127,
    "duration": 2.056,
    "user_input": [],
    "status": "not-afk",
    "app": "web",
    "events": []
},
{
    "timestamp": 1717335893.215,
    "duration": 10.267,
    "user_input": [
        {
            "from": "mouse",
            "data": {
                "type": "click",
                "button": "left"
            }
        },
        {
            "from": "keyboard",
            "type": "input",
            "data": "swift ui ctrl_l liebiao "
        },
        {
            "from": "keyboard",
            "data": {
                "type": "pressAndRelease",
                "key": "enter"
            }
        }
    ],
    "status": "not-afk",
    "app": "web",
    "events": []
},
{
    "timestamp": 1717335904.513,
    "duration": 0.0,
    "user_input": [],
    "status": "not-afk",
    "app": "web",
    "events": []
}]
\end{lstlisting}

\noindent\makebox[\linewidth]{\rule{\linewidth}{0.4pt}}\\
\textit{Processed Event}
\begin{lstlisting}[basicstyle=\ttfamily, breaklines=true]
[{
    "time": "1717378968.208",
    "event": "The user opens a new browser tab and navigates to the Google homepage."
},
{
    "time": "1717378971.255",
    "event": "The user switches to the 'Code.exe' application but does not perform any specific actions."
},
{
    "time": "1717378975.29",
    "event": "The user continues to remain on the 'Code.exe' application without performing any actions."
}]
\end{lstlisting}

\subsection{Proactive Agent's Prediction}
\noindent\makebox[\linewidth]{\rule{\linewidth}{0.4pt}}\\
\textit{Example 1}
\begin{lstlisting}[basicstyle=\ttfamily, breaklines=true]
{
    "observation": {
        "time": "1717378968.208",
        "event": "The user opens a new browser tab and navigates to the Google homepage."
    },
    "agent_response": [
        "Suggest checking the user's search history and providing personalized search recommendations."
    ],
    "task_status": false,
    "other_infomation": {
        "Purpose": "The user is opening a new browser tab and navigating to the Google homepage.",
        "Thoughts": "Based on the event, it seems the user is starting a new search or browsing session. They might need help with organizing their search or finding relevant information.",
        "Response": "Would you like me to review your search history and provide some personalized recommendations to help you find what you're looking for?"
    },
    "judgment": [
        false
    ]
}
\end{lstlisting}
\noindent\makebox[\linewidth]{\rule{\linewidth}{0.4pt}}\\
\textit{Example 2}
\begin{lstlisting}[basicstyle=\ttfamily, breaklines=true]
{
    "observation": {
        "time": "1717378971.255",
        "event": "The user switches to the 'Code.exe' application but does not perform any specific actions."
    },
    "agent_response": [
        "Suggest creating a new terminal or command prompt within the 'Code.exe' application to help the user with coding tasks."
    ],
    "task_status": false,
    "other_infomation": {
        "Purpose": "The user is switching to the 'Code.exe' application but didn't perform any specific actions.",
        "Thoughts": "Given the user's recent switch to the 'Code.exe' application, it's likely they're preparing to start coding or editing. However, they might need help with organizing their code, finding relevant documentation, or setting up their environment.",
        "Response": "I can help you create a new terminal or command prompt within the 'Code.exe' application. This will allow you to easily access coding tools and commands. Would you like me to set that up for you?"
    },
    "judgment": [
        false
    ]
}
\end{lstlisting}
\noindent\makebox[\linewidth]{\rule{\linewidth}{0.4pt}}\\
\textit{Example 3}
\begin{lstlisting}[basicstyle=\ttfamily, breaklines=true]
{
    "observation": {
        "time": "1717378975.29",
        "event": "The user continues to remain on the 'Code.exe' application without performing any actions."
    },
    "agent_response": [
        "Offer to provide coding assistance, such as suggesting online resources, tutorials, or even helping with writing code."
    ],
    "task_status": false,
    "other_infomation": {
        "Purpose": "The user is remaining on the 'Code.exe' application without performing any actions.",
        "Thoughts": "Given the user's recent actions, it seems they're hesitant to start coding or editing. They might be experiencing writer's block, struggling with a particular concept, or needing help with a specific coding task.",
        "Response": "If you're struggling with a coding task or need help with a specific concept, I'm here to assist. Would you like me to suggest some online resources, tutorials, or even help with writing code together?"
    },
    "judgment": [
        false
    ]
}
\end{lstlisting}

\subsection{Reward Model Judgment Example}
\noindent\makebox[\linewidth]{\rule{\linewidth}{0.4pt}}\\
\textit{Input Prompt}
\begin{lstlisting}[basicstyle=\ttfamily, breaklines=true]
{
    "Observations (Time Ascending)": [
        {
            "time": "1717342908.098",
            "event": "The user searched for 'remote working software' in the web browser and pressed 'Enter'."
        },
        {
            "time": "1717342914.314",
            "event": "A new tab titled 'new Tab' was opened in the web browser."
        },
        {
            "time": "1717342940.516",
            "event": "The user opened a search result with the title 'remote working software - search' on Bing."
        },
        {
            "time": "1717342956.012",
            "event": "The user switched to another tab in the web browser, interacting with multiple scroll actions."
        },
        {
            "time": "1717343061.447",
            "event": "The user resumed browsing on Bing search."
        },
        {
            "time": "1717343082.081",
            "event": "The user continued exploring search result pages in Bing with multiple scrolling actions and clicking on specific results."
        }
    ],
    "Proposed Task": null,
    "Instruction": "Now give your judgment. You should complete the reasoning process in the first person."
}
\end{lstlisting}

\noindent\makebox[\linewidth]{\rule{\linewidth}{0.4pt}}\\
\textit{Ouput Judgment}
\begin{lstlisting}[basicstyle=\ttfamily, breaklines=true]
{
    "thought": "I have been browsing the web and switching between different applications, and I haven't received any task proposal from the proactive assistant. Upon analyzing my current activities, I realize I've been searching for remote working software and exploring different search results on Bing. Given my current engagement and the fact that I haven't been interrupted or indicated a need for assistance, I believe there is no need for a task at this moment. The assistant's decision to not propose a task seems appropriate, as introducing a new task could potentially disrupt my workflow and cause unnecessary interruptions.",
    "judgment": "accepted"
}
\end{lstlisting}

\end{document}